# Personal Intelligence System UniLM: Hybrid On-Device Small Language Model and Server-Based Large Language Model for Malay Nusantara

*Azree Nazri[1,2,3], Olalekan Agbolade[2], Faisal Aziz[2]*

*azree@upm.edu.my*

[1]*Artificial Intelligence Soceity Malaysia*

[2]*Department of Computer Science, FSKTM, UPM, 43400, Serdang Selangor*

[3]*Instutute of Mathematical Research, UPM, 43400, Serdang Selangor*

**Abstract**

In contexts with limited computational and data resources, high-resource language models often prove inadequate, particularly when addressing the specific needs of Malay languages. This paper introduces a Personal Intelligence System designed to efficiently integrate both on-device and server-based models. The system incorporates SLiM-34M for on-device processing, optimized for low memory and power usage, and MANYAK-1.3B for server-based tasks, allowing for scalable, high-performance language processing. The models achieve significant results across various tasks, such as machine translation, question-answering, and translate IndoMMLU. Particularly noteworthy is SLiM-34M's ability to achieve a high improvement in accuracy compared to other LLMs while using 2 times fewer pre-training tokens. This work challenges the prevailing assumption that large-scale computational resources are necessary to build effective language models, contributing to the development of resource-efficient models for the Malay language with the unique orchestration between SLiM-34M and MANYAK-1.3B.

**Introduction**

Large language models (LLMs) have predominantly been developed using extensive datasets from high-resource languages, such as English, Chinese, and Spanish, leaving low-resource languages, particularly Malay, at a significant disadvantage (Minaee et al., 2024). A primary obstacle is the limited availability of high-quality textual data, which is essential for the effective training of LLMs. For many low-resource languages, this data is often fragmented, poorly standardized, or even non-existent. Additionally, the absence of critical tools, such as part-of-speech taggers or annotated corpora, further impedes the development of natural language processing (NLP) models. The Malay language exemplifies these challenges, as it remains underrepresented in digital resources and computational research. Moreover, the substantial computational resources required for training and deploying LLMs are often inaccessible, restricting the ability of local researchers and developers to utilize these technologies within their communities. These barriers underscore the need for innovative approaches to address data scarcity and resource limitations in the context of Malay and other low-resource languages. Developing language models that are both efficient and adaptable for refinement, fine-tuning, and deployment on limited hardware is increasingly important. While open-source models have made considerable progress in addressing language disparities, further efforts are required to build models that are not only computationally efficient but also locally relevant.

Given the significant linguistic diversity and the limited availability of digital resources for many languages, particularly in Malay, researchers have been actively exploring methods to make large language models (LLMs) more inclusive and effective for underrepresented languages. The challenges posed by the scarcity of well-structured data, coupled with limited computational resources, have necessitated the development of novel approaches aimed at overcoming these barriers.

One approach to addressing the challenges faced by low-resource languages involves the use of multilingual models and cross-lingual transfer learning. Models such as Multilingual BERT (mBERT) (Devlin et al., 2018), XLM-R (Conneau et al., 2019), and more recently Llama 3 (Dubey et al., 2024), have been trained on datasets spanning multiple languages, including several low-resource languages. These models capitalize on shared representations across different languages, enhancing their performance even when labelled data is sparse. However, their effectiveness for genuinely low-resource languages, such as Malay, remains limited due to the small volume of available training data.

In addition to general-purpose multilingual models (Qin et al., 2024), there has been a growing focus on developing specialised models that are tailored specifically to low-resource languages (Hedderich et al., 2021). These models frequently employ transfer learning techniques, whereby a model pre-trained on high-resource languages is fine-tuned using a smaller dataset from a low-resource language (Nekoto et al., 2020). This method not only enhances model performance but also significantly reduces the computational resources required, making it a more viable option in environments with constrained resources. This focus on resource-efficient strategies underscores the importance of improving accessibility to NLP technologies for languages with limited digital representation, such as Malay.

Despite notable advancements in large language models (LLMs), significant challenges persist in ensuring their effectiveness for low-resource languages like Malay. Issues such as linguistic bias, model interpretability, and the ethical implications of deploying these models in diverse cultural contexts remain pressing concerns. The lack of sufficient linguistic data exacerbates these challenges, raising questions about the fairness and inclusivity of such models.

This paper aims to address these gaps by developing a model that is both culturally sensitive and resource-efficient. The proposed approach focuses on building a low-resource model that can operate effectively despite the constraints of limited data and computational power. By prioritising efficiency and cultural awareness, the model seeks to overcome the limitations traditionally associated with LLMs in underrepresented languages, offering a more inclusive and ethically grounded solution for natural language processing in the Malay linguistic landscape.

*Small language model*

Zhang et al. (2024) introduced TinyLlama, a language model with 1.1 billion parameters pre-trained on a corpus of 1 trillion tokens. Despite its relatively compact size, TinyLlama utilises advanced techniques like FlashAttention to deliver strong performance across a range of tasks, outperforming many models within its parameter class. Building upon this foundation, the authors developed TinyLlama v1.1, which includes specialised versions of the model tailored to specific domains such as mathematics, code generation, and Chinese language processing. Through a multi-stage pretraining process, these domain-specific models demonstrate significant improvements in task-specific performance.

Parallel efforts to enhance model efficiency have led to the development of the OneBit framework by Xu et al. (2024), marking a key advancement in the quantization of large language models to 1-bit representations. This method significantly reduces both computational and memory demands, making it possible to deploy LLMs on devices with limited resources. Unlike conventional quantization techniques, which typically rely on 4-bit or 8-bit compression, OneBit achieves a remarkable compression ratio while maintaining a balance between size reduction and model accuracy across diverse tasks. This approach represents a critical step forward in enabling the use of LLMs in resource-constrained environments, including those encountered in Malay language processing.

In the pursuit of more resource-efficient training methods, Inheritune (Sanyal et al., 2024) presents an innovative approach to developing smaller language models by inheriting layers from a larger, pre-trained reference model, while relying on significantly reduced datasets. This methodology was demonstrated through the construction of a 1.5 billion parameter model, derived from a larger 3 billion parameter model. Remarkably, the smaller model was trained on a mere 1 billion tokens, representing just 0.1% of the dataset used for the original, larger model. Despite this reduction in training data, the resulting model achieved performance levels comparable to models trained on much larger datasets, underscoring its effectiveness in low-data environments.

This approach offers a promising solution for resource-constrained settings, such as the development of models for the Malay language, where both data and computational resources are limited. By leveraging Inheritune, smaller, yet effective, models can be developed without the need for extensive training data, making it an ideal method for addressing the challenges faced by low-resource languages.

Focusing on on-device processing, MobiLlama (Thawakar et al., 2024) is a 500 million parameter small language model (SLM) specifically optimised for use on resource-constrained devices. MobiLlama prioritises energy efficiency, reduced memory usage, and faster inference times, making it particularly suited for applications that require on-device processing. To achieve this, the researchers employed a parameter-sharing technique across the model's transformer layers, allowing the model to maintain high levels of accuracy while significantly reducing both training and deployment costs. MobiLlama was tested across nine benchmarks, consistently outperforming models in its class, especially in terms of efficiency on lower-end hardware.

In related research, Lepagnol et al. (2024) conducted a study on zero-shot text classification with small language models (SLMs), comparing models ranging from 77 million to 40 billion parameters across 15 diverse datasets. Their findings showed that smaller models could, in many cases, match or even surpass the performance of their larger counterparts, highlighting the efficiency of SLMs in particular tasks. The study also resulted in the creation of an open-source repository, providing valuable documentation of their methodologies.

Further exploring the capabilities of small language models, Scaria et al. (2024) investigated their ability to learn, retain, and unlearn noise patterns. The study involved models such as Olmo 1B, Qwen1.5 1.8B, Gemma 2B, and Phi2 2.7B, revealing that while these models can adapt to noise and even eliminate it, their performance varied significantly depending on the type of noise introduced, particularly at the character level. This research underscores the complexity of noise handling in SLMs and its impact on model performance.

This body of work collectively demonstrates how smaller, more efficient models such as MobiLlama can serve as powerful alternatives to larger models, particularly in settings where computational resources are limited, such as for Malay language applications. These studies highlight the potential of small language models in balancing performance and efficiency, especially in resource-constrained environments.

Zhu et al. (2024) introduced LLaVA-Phi, an efficient multi-modal assistant leveraging the Phi-2 small language model to enable multi-modal dialogue capabilities. Despite its relatively modest size of 2.7 billion parameters, LLaVA-Phi showcased impressive performance across a range of benchmarks, including visual comprehension and reasoning tasks. This model paves the way for novel applications in environments where real-time interaction is critical, demonstrating that smaller language models can handle sophisticated tasks while maintaining high resource efficiency.

In the domain of natural language processing, Brei et al. (2024) tackled the challenge of translating natural language into SPARQL queries using small language models. Models such as BART and M2M100 were tested on datasets like QALD and CoyPu, delivering solid results in SPARQL translation, while the performance of T5 models showed limitations in terms of accuracy.

In their pursuit of improved model efficiency, Song et al. (2024) focused on achieving sparse activation in small language models. By developing a new attribution metric, they overcame the challenges of existing sparse activation techniques, achieving an impressive 80% sparsification ratio with minimal accuracy loss, comparable to larger models.

Lastly, in the field of speech synthesis, Lemerle et al. (2024) introduced the Small-E model, a compact language model augmented with linear attention. Their research set a new standard in zero-shot voice cloning, highlighting the strong potential of small models in this specialised domain.

These developments, particularly in small language models like LLaVA-Phi and Small-E, illustrate the growing capacity of compact models to perform advanced tasks while ensuring resource efficiency, making them highly relevant for applications in low-resource settings, including those involving the Malay language.

*LLM for Malay Language*

Recent research on large language models (LLMs) for Malay languages has focused on several critical areas: the development of linguistic resources, the adaptation of models to better reflect the unique characteristics of the Malay language, and the enhancement of LLM performance for these underrepresented languages (Table 1). Efforts to create robust linguistic datasets have been central to this research, providing the necessary foundation for building models that are both accurate and culturally relevant. Additionally, researchers have been developing customised LLM architectures that are optimised for the specific linguistic features of Malay, ensuring that models perform more effectively than standard multilingual models. Lastly, there has been a concerted effort to improve the overall performance of these models, focusing on tasks such as machine translation, question-answering, and sentiment analysis to ensure their applicability across a range of natural language processing tasks.

**Table 1:** Multilingual Large Language Models (LLMs) Supporting Malay – Tokens Trained, Vocabulary Size, Parameters, and Language Inclusion

| Model | Tokens Trained (Trillion/Billion) | Vocabulary Size | Parameters (Million (M)/Billion (B)) | Malay Language Included |
|---|---|---|---|---|
| mBERT | Unknown | 110,000 | 110M | Yes |
| XLM-R | 2.5T | 250,002 | 270M | Yes |
| mT5 | 6T | 250,000 | 13B | Yes |
| mBART-50 | Unknown | 250,000 | 610M | Yes |
| M2M-100 | 7.5B | 128,000 | 12B | Yes |
| T5 Multilingual | 1T | 32,000 | 11B | Yes |
| ByT5 | Unknown | N/A | 13B | Yes |
| LLaMA (multilingual fine-tuned) | 1T | 32,000 | 7-65B | No (initially monolingual) |

*Languages*

The Malay language holds a prominent place in the Nusantara region, which encompasses the Malay Archipelago, including modern-day Malaysia, Indonesia, Brunei, Singapore, and parts of Thailand and the Philippines. It is a member of the Austronesian language family, one of the most widely dispersed language families in the world.

Malay has long served as a lingua franca in the Nusantara, facilitating trade, diplomacy, and cultural exchange across the region. Its widespread use is attributed to the Srivijaya Empire (7th to 13th century) and later the Malacca Sultanate (15th century), which established Malay as the language of administration and commerce.

There are several dialects and varieties of Malay spoken across the Nusantara, influenced by geography, culture, and history:

1. **Standard Malay (Bahasa Malaysia)**: The official language of Malaysia and Brunei, and a co-official language in Singapore. It is based on the Johor-Riau dialect.

2. **Indonesian (Bahasa Indonesia)**: The official language of Indonesia, it is a standardized form of Malay with distinct vocabulary and pronunciation, influenced by Dutch, Javanese, and other local languages.

3. **Bruneian Malay**: A variety spoken in Brunei, it differs slightly from the Malaysian standard but is mutually intelligible.

4. **Patani Malay**: Spoken in Southern Thailand, it has its own distinct features but remains close to the classical Malay language.

5. **Sabahan and Sarawakian Malay**: Variants spoken in the Malaysian Borneo states, which include unique local influences.

In the contemporary Nusantara, standard Malay continues to be a dominant language in formal education, government, media, and literature. In Malaysia and Brunei, Standard Malay serves as the national language, while in Indonesia, Bahasa Indonesia plays a unifying role in a linguistically diverse country with over 700 languages. The shared linguistic roots of Malay and Indonesian facilitate communication across national borders, although there are notable differences in vocabulary, pronunciation, and grammar.

In addition to its formal roles, Malay is also spoken as a mother tongue by various ethnic groups, including the Malay, Minangkabau, and Bugis, making it a living language in daily interactions across the region.

*Malaysia Malay Language*

The Malay language is the official language of Malaysia, spoken by the majority of the population and widely used in Brunei, Singapore, Indonesia, and parts of Thailand. It belongs to the Austronesian language family, specifically the Malayo-Polynesian branch, with around 30 million native speakers and an additional 200 million speakers across the region. The language has absorbed numerous loanwords, particularly from Arabic, due to historical Islamic influence, as well as from Sanskrit, Tamil, Portuguese, and English, reflecting its long-standing role as a regional lingua franca. Multiple dialects exist, with Bahasa Malaysia and Bahasa Indonesia being the most prominent standardized forms, though regional varieties, such as Kelantanese and Sarawak Malay, add further diversity to the language.

*Indonesia Language*

The Indonesian language, officially known as Bahasa Indonesia, is the national and official language of Indonesia. It is a standardized form of Malay, belonging to the Austronesian language family, specifically the Malayo-Polynesian branch. Bahasa Indonesia serves as a lingua franca across the archipelago, unifying a country with over 700 native languages. It is spoken by approximately 270 million people, including over 40 million who speak it as a first language.

Like Malay, Bahasa Indonesia has absorbed many loanwords from Arabic, Sanskrit, Portuguese, Dutch, and English, reflecting its historical interactions and colonial past. Despite being closely related to Bahasa Malaysia, the Indonesian variant has evolved separately, with distinct vocabulary and pronunciation differences.

*Brunie Language*

The Brunei Malay language, known locally as Bahasa Melayu Brunei, is the official language of Brunei. It is a variety of the Malay language and belongs to the Austronesian language family, specifically within the Malayo-Polynesian branch. While Standard Malay is used for formal communication, education, and government, Brunei Malay is widely spoken in everyday life and is considered the national vernacular.

Brunei Malay has around 220,000 native speakers and shares many similarities with the Johor-Riau dialect of Malay, which forms the basis of Bahasa Malaysia and Bahasa Indonesia. However, it has distinct differences in pronunciation, vocabulary, and grammar, making it unique to Brunei. The language has absorbed loanwords from Arabic, English, and Sanskrit, reflecting Brunei's Islamic heritage and colonial history under the British.

*Patani Malay*

Patani Malay, or Bahasa Melayu Patani, is a variety of the Malay language spoken primarily in the Patani region of southern Thailand, particularly in the provinces of Pattani, Yala, Narathiwat, and parts of Songkhla. This region is part of the historical Patani Sultanate, and the language is closely related to Kelantanese Malay, spoken across the border in Kelantan, Malaysia. Patani Malay belongs to the Austronesian language family, specifically the Malayo-Polynesian branch.

Although it shares many similarities with Standard Malay and Kelantanese Malay, Patani Malay has distinct linguistic features, especially in its pronunciation, vocabulary, and idiomatic expressions. Thai is the official language in Thailand, so many Patani Malay speakers are bilingual, using Thai for official and educational purposes while reserving Patani Malay for informal, everyday conversations.

Given its historical and cultural context, Patani Malay has also absorbed influences from Arabic (due to the region's Islamic heritage) and Thai, reflecting the sociopolitical landscape in southern Thailand. However, unlike in Malaysia or Brunei, where Malay enjoys official status, Patani Malay is not officially recognized in Thailand, and efforts to preserve it are often challenged by the dominance of Thai in official and educational domains.

*Sabanans and Sarawakians Malay*

Sabahan Malay and Sarawakian Malay are regional dialects of Malay spoken in the East Malaysian states of Sabah and Sarawak on the island of Borneo. While both dialects are closely related to Standard Malay, they have distinct linguistic features influenced by the local cultures, indigenous languages, and geographical context of their respective regions.

Sabahan Malay, also known as Sabah Malay or Sabahan Creole Malay, is spoken primarily in Sabah and is widely used as a lingua franca across the state's ethnically diverse population. It has significant influences from Bajau, Dusun, and Kadazan languages, as well as other local indigenous languages. Sabahan Malay has its own unique vocabulary, pronunciation, and sentence structures that set it apart from Standard Malay. For instance, some words and phrases are borrowed or adapted from local languages, giving it a distinct Sabahan flavor. It is predominantly used in informal settings, with Standard Malay and English used in formal and educational contexts.

Sarawakian Malay, or Sarawak Malay, is spoken mainly in Sarawak and is distinct from Standard Malay due to its strong influence from Iban, Bidayuh, and other indigenous languages. Sarawak Malay often simplifies or shortens certain words and has a more relaxed pronunciation compared to Standard Malay. For instance, words in Sarawakian Malay may be contracted or abbreviated, creating a more colloquial and local tone. It is commonly spoken in day-to-day interactions across the state and serves as an important marker of Sarawakian identity. Like Sabahan Malay, Sarawakian Malay is primarily used in informal settings, while Standard Malay is the medium for formal communication.

Both Sabahan and Sarawakian Malay are heavily influenced by the indigenous languages of East Malaysia, resulting in vocabularies that reflect the local cultural context. For example, words related to nature, local customs, and food are often borrowed from indigenous languages. This unique blend of Malay and local languages distinguishes Sabahan and Sarawakian dialects from their Peninsular Malaysian counterparts.

**Table 2:** Dataset used for pre-training.

| Source | Standard Malay | Indonesian | Brunei | Patani Malay | Sabahan Creole Malay | Sarawakian Malays |
|---|---|---|---|---|---|---|
| Wikipedia | x | x | x | x | x | X |
| Study Article | x | x | X | | | |
| Malaysia/Indonesia/Brunie/Thailand Government Public documents | x | x | X | | | |
| Malaysia/Indonesia/Brunie/Thailand public articles | x | x | x | x | x | X |
| Malaysia/Indonesia/Brunie/Thailand public journals | x | x | X | | | |
| Malaysia/Indonesia/Brunie/Thailand related public research papers | x | x | x | | | |
| YoutubeSubtitles | x | x | x | x | x | X |
| OpenSubtitles | x | x | x | x | x | X |
| Total Malay Nusantara Tokens | 150B (large)/2.4B(small) | | | | | |
| Total Malay Nusantara Vocabulary | 299K (large)/15K (small) | | | | | |

*Dataset*

Table 2 provides a comprehensive overview of the datasets used for pre-training Malay Nusantara language models, covering various dialects including Standard Malay, Indonesian, Brunei Malay, Patani Malay, Sabahan Creole Malay, and Sarawakian Malay. The sources of data for pre-training range from publicly available resources like Wikipedia and OpenSubtitles to government documents, public articles, and research papers from Malaysia, Indonesia, Brunei, Thailand, and surrounding regions. These datasets are essential for developing models that can understand and process the linguistic diversity of the Malay Nusantara languages. In terms of scale, the total tokens used for pre-training the large model amount to 150 billion tokens, while the small model uses 2.4 billion tokens. The vocabulary size also differs significantly between the large (299K) and small models (15K). This illustrates the breadth of coverage in the data, which ensures that the model can handle a wide variety of Malay dialects and contexts efficiently. This diverse and rich dataset makes the models more robust for various applications, including language translation, text generation, and sentiment analysis across these different dialects.

*Pre-training*

*Personal Intelligence System (PIS) architcture*

Figure 1 represents integration of SLiM-34M on-device and MANYAK-1.3B server-based models so-called Personal Intelligence System (PIS). The architecture is designed to balance computational efficiency, privacy, and performance, leveraging both local device models and cloud-based resources.

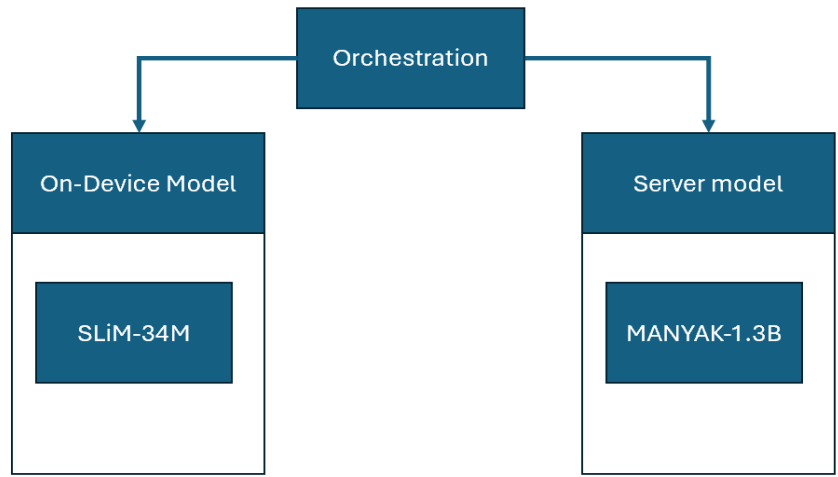

**Figure 1 Architecture Overview**: Orchestration of SLiM-34M On-Device Model and MANYAK-1.3B Server Model

*SLiM-34M On-Device Model*

SLiM-34M on-device model handles tasks such as language and image processing directly on on-devices, powered by the device's CPU, GPU, Neural Engine, and Secure Enclave. These models are optimized for tasks requiring low latency and enhanced privacy, ensuring that sensitive user data stays local. Both the on-device and server models utilize Grouped-Query-Attention (GQA), enhancing the efficiency of the attention mechanism, which is crucial for managing the computational demands of modern deep learning models.

To minimize memory and inference costs, shared input and output vocab embedding tables are used across the models. For on-device inference, low-bit palletization is employed — a crucial optimization technique that meets the stringent memory, power, and performance requirements for mobile devices. This technique uses a combination of 2-bit and 4-bit configurations for weight storage, averaging 3.5 bits per weight, which ensures the compressed models achieve comparable accuracy to their uncompressed counterparts. This allows Apple to maintain high-quality models on resource-constrained hardware.

*MANYAK-1.3B Server Model*

Manyak-1.3B server model deployed on Cloud Compute handle more complex, resource-intensive tasks. These models utilize cloud's extensive computational infrastructure, including the ML stack, Private Cloud Extensions, and Private Cloud Compute OS, allowing for high-performance processing when needed. Manyak-1.3B server-based model is capable of dynamically loading adapter models, typically requiring tens of megabytes for rank 16 adapters, to fine-tune and specialize on-the-fly based on task requirements.

*Orchestration*

The Orchestration layer centrally manages which tasks are processed locally on the device and which are offloaded to the cloud. This system ensures smooth user experiences by dynamically allocating resources, guaranteeing the operating system's responsiveness, and maintaining a balance between computational load and efficiency.

*Training Details*

The SLiM-34M and Manyak-1.3B model was trained using 32 instances of AWS EC2 p4d.24xlarge, a highly optimized cloud computing infrastructure suited for machine learning and deep learning tasks. The training utilized Nvidia A100 40GB GPUs, which are specifically designed to accelerate AI workloads, delivering high computational performance for model training. The training process spanned 16 days.

### *Malay Nusantara Large Language Model (MANYAK-1.3B)*

Figure 2 provided illustrates the process involved in developing a Malay Nusantara Large Language Model (LLM) using Megatron GPT architecture, focusing on a bilingual tokenizer strategy. The key steps in this process are as follows:

- **Malay Nusantara (MS) JSONL & GPT2 BPE Pretrained Tokenizer:** This phase utilizes the Malay Nusantara dataset (Standard Malay, Indonesia, Brunie, Patani Malay, Sabahan and Sarawakian Malay) in the JSONL format and combines it with the GPT-2 Byte Pair Encoding (BPE) tokenizer to initialize tokenization for the Malay language data.
- **MS BPE Tokenizer:** A specialized tokenizer is generated for Malay Nusantara (MS) using BPE, ensuring the language data is efficiently tokenized for training.
- **English GPT-1.3B BPE Tokenizer:** This component shows how the English model is integrated into the process. The English GPT-1.3B model is trained with its BPE tokenizer.
- **Merged En/MS BPE Tokenizer:** The English and Malay Nusantara BPE tokenizers are merged, creating a joint tokenization system that supports both languages, facilitating bilingual capabilities.
- **Megatron GPT-1.3B MS:** The Megatron GPT-1.3B model is fine-tuned specifically for the Malay Nusantara language (MS). This step involves adapting the English pre-trained model to Malay Nusantara language tasks.
- **Megatron GPT-1.3B Model with Extended Embedding Layer:** The embedding layer of the Megatron GPT-1.3B model is expanded to accommodate the newly merged vocabulary (English + Malay Nusantara). This ensures the model can effectively process and represent both languages.
- **Malay LLM Model:** The final output is a dedicated Malay Nusantara Large Language Model, built upon the multilingual tokenizer and fine-tuned model structure. It represents a powerful NLP tool tailored to the Malay Nusantara language using modern transformer-based techniques.

The entire process combines techniques like byte pair encoding (BPE), multilingual tokenization, and transfer learning (from the English Megatron GPT model) to create a robust language model capable of understanding and generating text in Malay Nusantara. This approach maximizes the use of pre-existing resources while optimizing the model for local linguistic nuances.

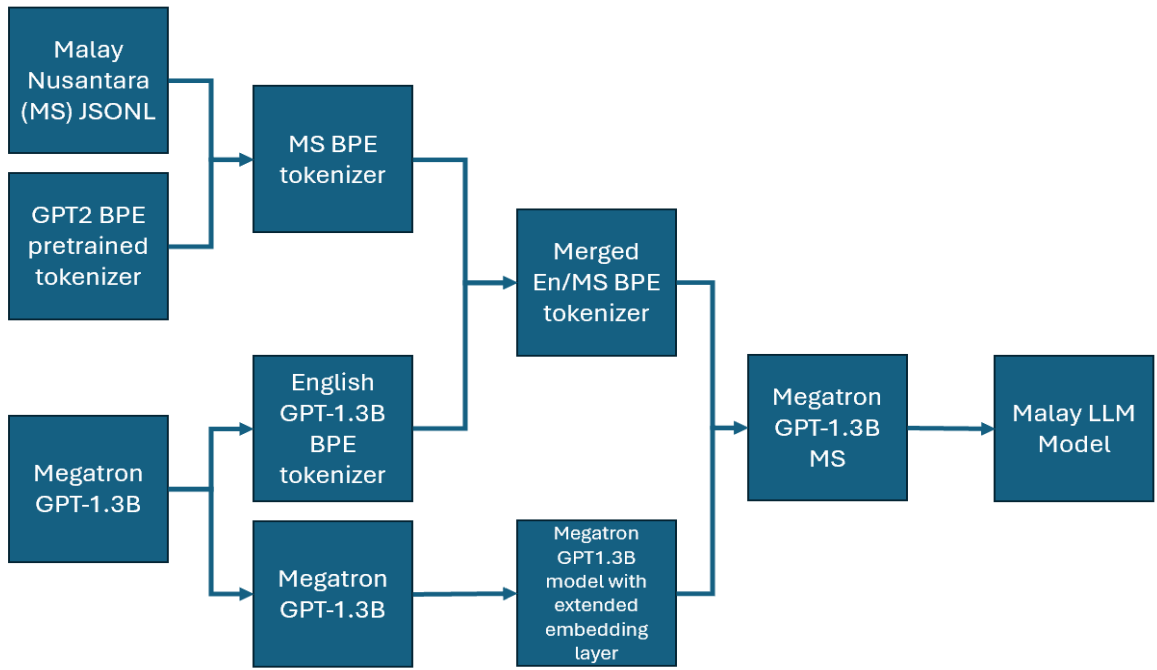

**Figure 2:** Workflow for training a localized Malay LLM

*Malay Nusantara Small Language Model (SLiM-34M)*

The architecture of SLiM-34M adopts a decoder-only transformer-based model, aligning with the principles followed by state-of-the-art small language models (SLMs) as shown in Table 3. Specifically, the following key innovations are incorporated: (1) Learnable bias parameters are excluded from all fully-connected layers, optimizing efficiency, (2) RMSNorm is applied for pre-normalization, ensuring smoother training, and rotatory positional embedding (ROPE) is used to encode positional information, enhancing the model's contextual understanding, (3) Grouped Query Attention (GQA) is utilized instead of the conventional multi-head attention (MHA), reducing the number of attention heads while maintaining performance, (4) The Feed Forward Network (FFN) is replaced with SwiGLU FFN, which improves activation functions for better model learning, (5) Flash Attention is employed to compute scaled dot-product attention, significantly improving computational speed, and (6) The model uses the same tokenizer as LLama, ensuring efficient tokenization and language comprehension.

Table 3: SLiM-34M architecture and hyperparameter

| Hyperparameter | Value |
|---|---|
| Total Parameters | 0.422B |
| Hidden Size | 2048 |
| Intermediate Size (SwiGLU FFN) | 5632 |
| Number of Attention Heads | Grouped Query Attention (32 heads) |
| Number of Hidden Layers | 8 |
| RMSNorm | $1 \times 10^{-5T}$ |
| Max Seq Length | 2048 |
| Vocab Size | 61788 |
| Flash Attention | Yes |
| Learnable Bias Parameter (fully-connected layers) | No |

**Evaluation**

*Evaluation tool*

We evaluate models with the Bhasa and IndoMMLU datasets. To evaluate models using the Bhasa benchmark, we select key evaluation metrics such as F1 scores for tasks like QA (Question Answering), Sentiment Analysis, and Toxicity Detection. For machine translation tasks, we use ChrF++ for evaluating English to Standard Malay/Indonesian/Patani Malay/Sabahan/Sarawakian and Standard Malay/Indonesian/Patani Malay/Sabahan/Sarawakian to English translations. ROUGE-L is applied for text summarization tasks, while Accuracy (Acc) is measured for Natural Language Inference (NLI) and Causal Reasoning. These diverse tasks ensure that the models are evaluated comprehensively across multiple dimensions, making it ideal for assessing performance in low-resource languages like Malay, Indonesian, Patani Malay, Sabahan and Sarawakian Malay.

Figure 3 represents the IndoMMLU dataset, categorized according to different education levels and subject matter, segmented into Primary School (SD), Junior High School (SMP), Senior High School (SMA), and University Entrance Tests. Each of these educational tiers is further divided into thematic subjects: Local, Indo (Indonesian language), Humanities (Hum), Social Sciences (Social), and STEM (Science, Technology, Engineering, and Mathematics).

- **Primary School (SD) accounts for 30%** of the dataset. It focuses on basic education, represented by subjects like Local knowledge, Indonesian, Humanities, Social Sciences, and STEM.

- **Junior High School (SMP) accounts for 24%**, emphasizing more specialized knowledge with a similar division of subjects.

- **Senior High School (SMA) accounts for 32%**, a significant portion of the dataset, including subjects relevant for higher education preparatory knowledge.

- **University Entrance Tests account for 14%**, targeting advanced subjects designed to prepare students for university-level studies.

The diagram demonstrates the structured breakdown of educational data in the IndoMMLU, which is critical for multilingual natural language processing (NLP) tasks such as question answering, reading comprehension, and other educational benchmarks tailored for the Indonesian language and context.

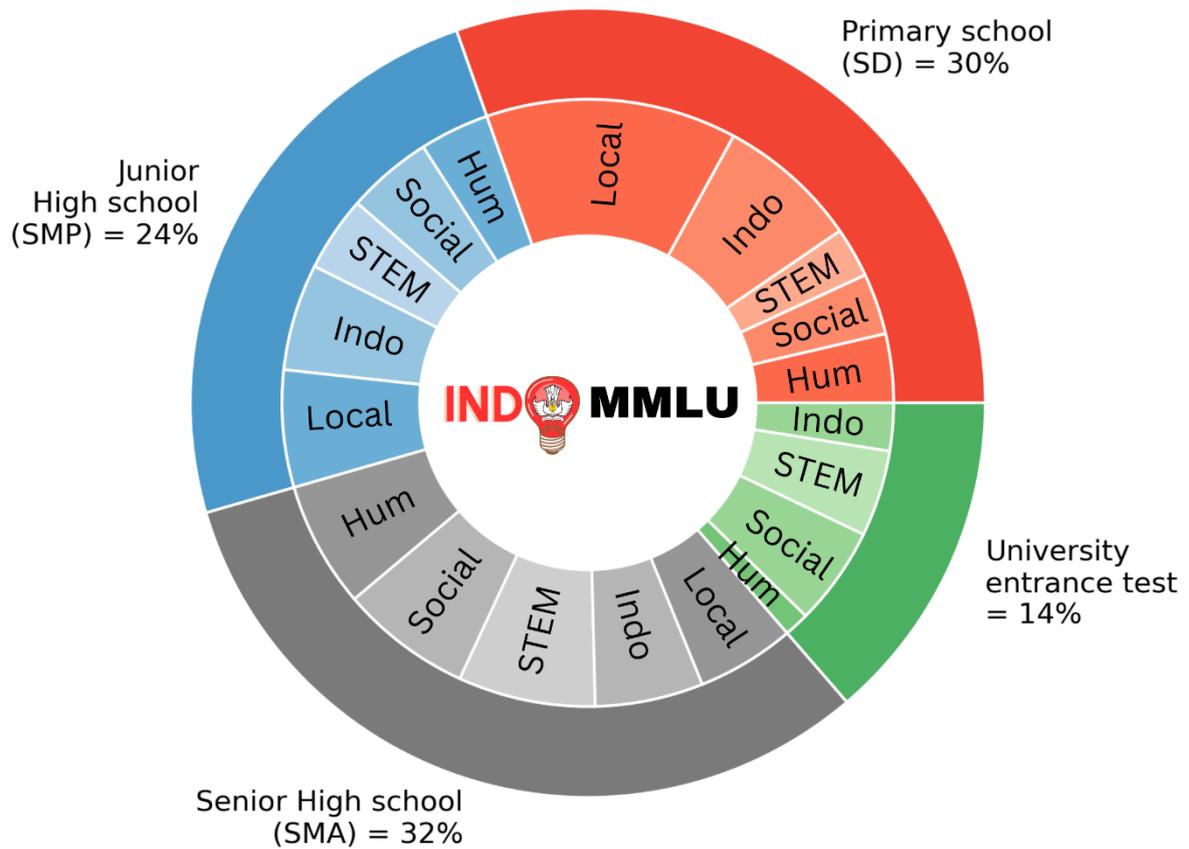

**Figure 3:** Distribution of subject areas and education levels in IndoMMLU. "Hum", "Social", "Indo", and "Local" refer to Humanities, Social Science, Indonesian Language, and Local Languages and Cultures, respectively.

To compare the Bhasa and IndoMMLU datasets to Malay Nusantara (Standard Malay, Brunei, Patani Malay, Sabahan, and Sarawakian Malay), the datasets in Indonesia are first translated into English, then into Malay Nusantara. The translation with low perplexity, high BLUE, and ROUGH is then chosen as shown in Figure 4.

**Minangkabau culture subject, Mid-term exam, class 7 (SMP)**

| Terjadinya hubungan induak bako dengan anak pisang, karena adanya .... | *Induak bako* and *anak pisang* is a relationship in Minangkabau family because of .... |
|---|---|
| **A. Perkawinan**<br>B. Satu suku<br>C. Satu nagari<br>D. Kaum | **A. Marriage**<br>B. One tribe relationship<br>C. One village relationship<br>D. One sub-tribe relationship |

**Minangkabau culture subject, Mid-term exam, class 6 (SD)**

**Indonesia**          **English**

| "Induak bako" dan "anak pisang" adalah hubungan dalam keluarga Minangkabau kerana |
|---|
| (a) perkahwinan |
| (b) hubungan satu suku |
| (c) hubungan satu kampung |
| (d) hubnugan satu sub suku |

**Standard Malay**

**Figure 4:** Translation from Indonesia to English into Standard Malay with low perplexity, high BLUE, and rough

*Model selection*

To evaluate models, we selected open-source models, comparing their performance across the tasks described in Section 6.3. Our evaluation contrasts the capabilities of small, multilingual, and larger models against our SLiM-34M model. The models chosen include MiniLM (22M parameters) and DistilBERT (66M parameters) as smaller models, while mBERT (110M parameters) and XLMR-Base (125M parameters) represent multilingual models. Additionally, DistilGPT (82M parameters) and GPT-Neo (125M parameters) were included for further comparison. This range of models ensures a comprehensive analysis across different model sizes and architectures.

**Results**

*Malay-translated Bhasa evaluation*

Table 4 evaluates the performance of several models on tasks such as question-answering (QA), sentiment analysis, toxicity detection, machine translation, summarization, natural language inference (NLI), and causal reasoning, comparing their accuracy and effectiveness across various metrics.

The mBERT (Multilingual BERT) model performs relatively poorly across most tasks. It scores 23.76 on QA, showing a limited ability to understand and generate appropriate answers. In sentiment analysis, mBERT achieves 31.46, which is mediocre compared to more recent models, and it performs even worse in toxicity detection with an F1 score of 11.84. In machine translation, both for English-to-Malay and Malay-to-English, mBERT struggles, achieving 21.48 and 21.9, respectively. The model also underperforms in summarization, with a ROUGE-L score of 11.28, and its NLI accuracy is only 15.45, reflecting difficulty in complex reasoning tasks. Similarly, for causal reasoning, mBERT performs poorly with an accuracy of 12.04, indicating limited understanding of cause-effect relationships.

XLM-R (Cross-lingual Language Model) shows some improvement over mBERT. It achieves a slightly higher score in QA (27.71) but still lags behind the more advanced models. For sentiment analysis, XLM-R scores 27.85, which is lower than expected, and its performance in toxicity detection (F1 score of 12.43) is also suboptimal. In translation tasks, XLM-R scores 24.42 for English-to-Malay and 25.48 for Malay-to-English, indicating moderate translation abilities. Its summarization ability, with a ROUGE-L score of 14.73, is better than mBERT but still below the top-performing models. In NLI, XLM-R has an accuracy of 11.4%, and in causal reasoning, it shows some improvement over mBERT, with an accuracy of 14.23.

The mT5 model demonstrates modest improvements across several tasks. It scores 25.26 on QA, which is better than mBERT but still not competitive with more recent models. In sentiment analysis, it achieves 29.25, and in toxicity detection, it performs better with an F1 score of 19.98. For translation, mT5 scores 26.53 (English-to-Malay) and 26.73 (Malay-to-English), showing more competent translation skills. Its ROUGE-L score of 18.7 suggests better summarization capabilities, while its NLI performance (accuracy of 13.15) remains in the lower range. mT5 shows some improvement in causal reasoning, with an accuracy score of 15.55.

The mBART-50 model does not perform as well as mT5, particularly in sentiment analysis, with a score of 23.72, and it struggles with toxicity detection, scoring 11.93. Its translation abilities are weak, with scores of 22.33 and 21.68 for English-to-Malay and Malay-to-English, respectively. In summarization, mBART-50 achieves a ROUGE-L score of 11.43, and for NLI and causal reasoning, it shows slight improvements over mBERT, with accuracy scores of 15.96 and 11.18, respectively.

M2M-100 performs notably better in translation tasks, scoring 29.82 for English-to-Malay and 25.79 for Malay-to-English, demonstrating strong capabilities in multilingual translation. It also performs well in sentiment analysis, scoring 32.85, and in toxicity detection, with an F1 score of 21.73. Its summarization ability is decent, with a ROUGE-L score of 15.03, and it shows moderate performance in NLI and causal reasoning, with accuracy scores of 14.88 and 19.96, respectively.

The T5 Multilingual model, while not excelling in translation tasks, performs reasonably well in sentiment analysis with a score of 37.09 and also shows solid performance in summarization, achieving 22.27 in ROUGE-L. However, it performs poorly in QA (21.92) and other reasoning tasks, with an accuracy of 12.68% in NLI and 11.25% in causal reasoning, suggesting limitations in reasoning and understanding.

ByT5 achieves balanced performance across tasks. It performs well in toxicity detection (26.77) and moderately in sentiment analysis (31.35). Its translation scores of 29.49 for English-to-Malay and 23.95 for Malay-to-English indicate decent bilingual capabilities. In summarization, ByT5 scores 22.27 (ROUGE-L), and in NLI and causal reasoning, it shows improvements over other models, achieving accuracy scores of 20.27 and 29.45, respectively.

**Table 4:** Performance Comparison of Language Models Across Various Tasks in Malay Language

| Model | QA (F1) | Sentiment (F1) | Toxicity (F1) | Eng>Malay (ChrF++) | Malay>English (ChrF++) | Summary (ROUGE-L) | NLI (Acc) | Causal (Acc) |
|---|---|---|---|---|---|---|---|---|
| *mBERT* | 23.76 | 31.46 | 11.84 | 21.48 | 21.9 | 11.28 | 15.45 | 12.04 |
| *XLM-R* | 27.71 | 27.85 | 12.43 | 24.42 | 25.48 | 14.73 | 11.4 | 14.23 |
| *mT5* | 25.26 | 29.25 | 19.98 | 26.53 | 26.73 | 18.7 | 13.15 | 15.55 |
| *mBART-50* | 21.36 | 23.72 | 11.93 | 22.33 | 21.68 | 11.43 | 15.96 | 11.18 |
| *M2M-100* | 28.44 | 32.85 | 21.73 | 29.82 | 25.79 | 15.03 | 14.88 | 19.96 |
| *T5 Multilingual* | 21.92 | 37.09 | 20.56 | 31.47 | 27.32 | 12.6 | 12.68 | 11.25 |
| *ByT5* | 29.95 | 31.35 | 26.77 | 29.49 | 23.95 | 22.27 | 20.27 | 29.45 |
| *SLiM-34M* | 27.56 | 28.06 | 22.28 | 25.78 | 23.97 | 25.31 | 25.57 | 29.01 |
| *MANYAK-1.3B* | 75.75 | 95.18 | 63.22 | 65.71 | 69.66 | 40.18 | 76.36 | 90.86 |

SLiM-34M, a smaller model, performs surprisingly well across tasks, especially in sentiment analysis, with an F1 score of 28.06. It also excels in machine translation, scoring 25.78 for English-to-Malay and 23.97 for Malay-to-English. The model achieves a summarization ROUGE-L score of 25.31, indicating strong summarization capabilities for its size. Its NLI performance is noteworthy, with an accuracy score of 25.57, and in causal reasoning, SLiM-34M achieves 29.01%, making it highly competitive in reasoning tasks.

MANYAK-1.3B dominates across all tasks, performing significantly better than the other models. It scores an impressive 75.75 in QA and 95.18 in sentiment analysis, far surpassing other models. It excels in toxicity detection, achieving 63.22, and in translation tasks, it scores 65.71 for English-to-Malay and 69.66 for Malay-to-English, demonstrating exceptional bilingual capabilities. MANYAK-1.3B also achieves the highest score in summarization (40.18, ROUGE-L), and it dominates in reasoning tasks, with an accuracy of 76.36% in NLI and 90.86% in causal reasoning.

*Malay-translated IndoMMLU*

As shwon in Table 5, in comparison, the Random baseline model shows significantly lower results in all categories, further emphasizing the effectiveness of the trained models.

For mBART-50, the token count remains unknown, but its performance is generally lower compared to the other models, particularly in STEM (20.2) and Humanities (24.7). The model may benefit from further refinements, given its lower scores in several categories.

M2M-100, which was trained on 7.5 billion tokens, shows a balanced performance across the subjects, with a standout score of 25.79 in Local Culture and 29.82 in translating English to Malay. This model's broader training on multiple languages could explain its versatility across disciplines.

T5 Multilingual, trained on 1 trillion tokens, also performs moderately well, particularly in STEM (24.5) and Local Culture (27.5), though its performance in Humanities (24.8) and Social Science (22.6) suggests room for improvement.

ByT5, despite an unknown number of training tokens, performs similarly across the board, with scores in the 21.7–24.7 range for all subjects. This consistent performance might indicate that it is effective at handling tasks, even with less transparency around its training data.

SLiM-34M, trained on only 34 million tokens, performs relatively well for a smaller model, achieving a score of 26.2 in Humanities and 28.4 in Local Culture. Its higher scores in these areas suggest it is well-optimized for tasks requiring cultural understanding, despite its smaller size.

Lastly, MANYAK-1.3B, trained on 1.3 billion tokens, achieves the highest overall performance across all categories, with particularly high scores in Local Culture (91.45) and Social Science (68.51). This model's large training corpus and superior architecture likely contribute to its excellent performance in a wide range of tasks.

Table 5: Performance Comparison of Malay-Translated IndoMMLU Across Different Language Models

| Model | Tokens Trained (Trillion(T)/Billion(B)/Million(M)) | STEM | Social Science | Humanities | Local L. Culture |
|---|---|---|---|---|---|
| *mBERT* | Unknown | 20.9 | 20.5 | 24.9 | 26.7 |
| *XLM-R* | 2.5T | 27.4 | 28.9 | | |
| *mT5* | 6T | 34.1 | 36.8 | 39.9 | 40.5 |
| *mBART-50* | Unknown | 20.2 | 22.1 | 24.7 | 24.7 |
| *M2M-100* | 7.5B | 23.8 | 22.8 | 24.7 | 25.6 |
| *T5 Multilingual* | 1T | 24.5 | 22.6 | 24.8 | 27.5 |
| *ByT5* | Unknown | 21.7 | 21.7 | 24.7 | 24.2 |
| *SLiM-34M* | 34M | 23.1 | 24.5 | 26.2 | 28.4 |
| *MANYAK-1.3B* | 1.3B | 68.51 | 91.45 | 55.24 | 64.7 |
| *Random* | - | 21.9 | 23.4 | 23.5 | 26.6 |

**Discussion**

In contexts where computational and data resources are constrained, the limitations of high-resource language models become evident, particularly in addressing the specific needs of languages like Malay. This study presents the development of a Personal Intelligence System that bridges these challenges by incorporating both on-device and server-based models. The system integrates SLiM-34M, an on-device model optimized for low memory and power consumption, and MANYAK-1.3B, a server-based model designed for high-performance, scalable tasks. This dual-architecture setup allows for the efficient distribution of computational workloads across a variety of language processing tasks.

The models were evaluated on tasks such as machine translation, question-answering, and the Malay-Translated IndoMMLU benchmark. SLiM-34M demonstrates its value by achieving a notable high improvement in accuracy compared to other large language models, while utilizing only half the amount of pre-training tokens. This is significant because it challenges the assumption that high-performance models require large-scale computational resources, instead showing that a resource-efficient model can deliver competitive results.

MANYAK-1.3B, the server-based model, excels in handling complex, resource-intensive tasks, complementing SLiM-34M's on-device capabilities. The orchestration between these two models exemplifies the potential of a hybrid system that efficiently manages local device constraints while leveraging the power of larger, server-based models for more demanding tasks. This study contributes to the development of resource-efficient models tailored for the Malay language, offering an effective solution for environments where computational resources are limited but high performance is still required. The results pave the way for future innovations in language model architectures that prioritize efficiency without sacrificing accuracy or scalability.

**Conclusion**

In conclusion, this study demonstrates the potential of a hybrid Personal Intelligence System that integrates on-device and server-based models to address the specific needs of Malay language processing in resource-constrained environments. By combining the lightweight, efficient SLiM-34M model for on-device tasks with the scalable, high-performance MANYAK-1.3B server-based model, the system achieves significant results in key tasks such as machine translation, question-answering, and the Malay-Translated IndoMMLU benchmark. Notably, SLiM-34M achieves a high improvement in accuracy over comparable large language models, while utilizing half the pre-training tokens, challenging the assumption that large-scale computational resources are necessary for effective language modeling.

This research contributes to the growing body of work on resource-efficient language models, providing a pathway for future developments in language processing that do not require significant computational infrastructure. The unique orchestration of SLiM-34M and MANYAK-1.3B illustrates that even in limited-resource settings, it is possible to develop high-performing language models that can meet the demands of a wide range of tasks, ultimately advancing the field of language modeling for underrepresented languages like Malay.